\documentclass{article}

\PassOptionsToPackage{square,sort,comma,numbers}{natbib}
\usepackage[final]{nips_2017}

\usepackage[utf8]{inputenc} 
\usepackage[T1]{fontenc}    
\usepackage{hyperref}       
\usepackage{url}            
\usepackage{booktabs}       
\usepackage{nicefrac}       
\usepackage{microtype}      

\usepackage{hyperref}

\usepackage{amsmath,amsthm,amssymb,amsfonts}       
\usepackage[pdftex]{graphicx}
\usepackage{subcaption}
\usepackage{xfrac}
\usepackage{enumitem}
\usepackage{makecell}
\setlist{itemsep=0pt,topsep=0.1pt}

\newcommand{\R}{\mathbb{R}}

\newcommand{\Nm}{\mathcal{N}}

\newcommand{\rbr}[1]{\left(#1\right)}
\newcommand{\sbr}[1]{\left[#1\right]}
\newcommand{\cbr}[1]{\left\{#1\right\}}
\newcommand{\nbr}[1]{\left\|#1\right\|}
\newcommand{\abr}[1]{\left|#1\right|}

\DeclareMathOperator*{\argmin}{arg\,min}

\newcommand\numberthis{\addtocounter{equation}{1}\tag{\theequation}}

\title{Novel Prediction Techniques Based\\ on Clusterwise Linear Regression}

\author{
  Igor Gitman \\
  Machine Learning Department \\
  Carnegie Mellon University\\
  \texttt{igitman@andrew.cmu.edu} \\
  \And
  Jieshi Chen \\
  Auton Lab \\
  Carnegie Mellon University\\
  \texttt{jieshic@andrew.cmu.edu} \\
  \And
  Eric Lei \\
  Machine Learning Department \\
  Carnegie Mellon University\\
  \texttt{elei@andrew.cmu.edu} \\
  \And
  Artur Dubrawski \\
  Auton Lab \\
  Carnegie Mellon University\\
  \texttt{awd@cs.cmu.edu} \\
}

\begin{document}

\maketitle

\begin{abstract}
In this paper we explore different regression models based on Clusterwise Linear Regression (CLR). CLR aims to find the partition of the data into $k$ clusters, such that linear regressions fitted to each of the clusters minimize overall mean squared error on the whole data. The main obstacle preventing to use found regression models for prediction on the unseen test points is the absence of a reasonable way to obtain CLR cluster labels when the values of target variable are unknown. In this paper we propose two novel approaches on how to solve this problem. The first approach, \textit{predictive CLR} builds a separate classification model to predict test CLR labels. The second approach, \textit{constrained CLR} utilizes a set of user-specified constraints that enforce certain points to go to the same clusters. Assuming the constraint values are known for the test points, they can be directly used to assign CLR labels. We evaluate these two approaches on three UCI ML datasets as well as on a large corpus of health insurance claims. We show that both of the proposed algorithms significantly improve over the known CLR-based regression methods. Moreover, predictive CLR consistently outperforms linear regression and random forest, and shows comparable performance to support vector regression on UCI ML datasets. The constrained CLR approach achieves the best performance on the health insurance dataset, while enjoying only $\approx 20$ times increased computational time over linear regression.
\end{abstract}

\section{Introduction}
Clusterwise Linear Regression (CLR)~\cite{spath1979algorithm} is an effective technique for analyzing data that is assumed to be generated from a mixture of linear regression models. Formally, given a dataset with $n$ points and $d$ features $X \in \R^{n \times d}, Y \in \R^n$, CLR aims to find partition of the data into $k$ disjoint clusters, such that sum of squared errors of linear regression models inside each cluster is minimized:
\begin{align*}
    &\min_{C_i, w_i, b_i}\sum_{i=1}^k\sum_{j \in C_i}\rbr{y_j - x_j^Tw_i - b_i}^2 \\
    &\text{s.t.}\ \cup_{i=1}^k C_i = \{1,\dots,n\}, C_i \cap C_j = \varnothing\ \text{for}\ i \neq j \numberthis \label{eq:clr}.
\end{align*}

This problem is known to be NP-hard~\cite{megiddo1982complexity}; however many approximate algorithms have been proposed to solve it e.g.~\cite{spath1979algorithm},~\cite{desarbo1988maximum},~\cite{desarbo1989simulated},~\cite{bagirov2015algorithm}.

\begin{figure}[t]
    \centering
    \begin{subfigure}{0.45\textwidth}
        \centering
        \includegraphics[width=1.0\textwidth]{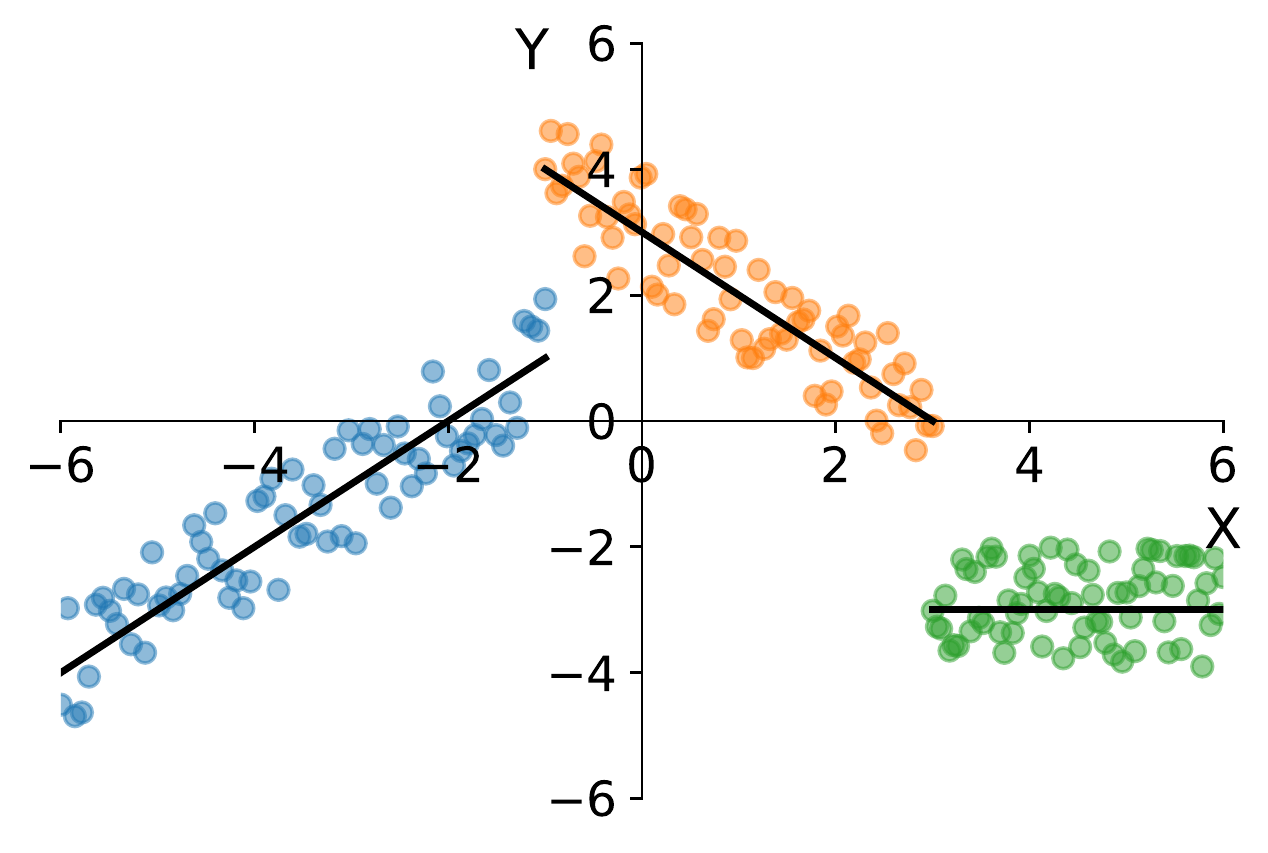}
        \caption{}
    \end{subfigure}
    \begin{subfigure}{0.45\textwidth}
        \centering
        \includegraphics[width=1.0\textwidth]{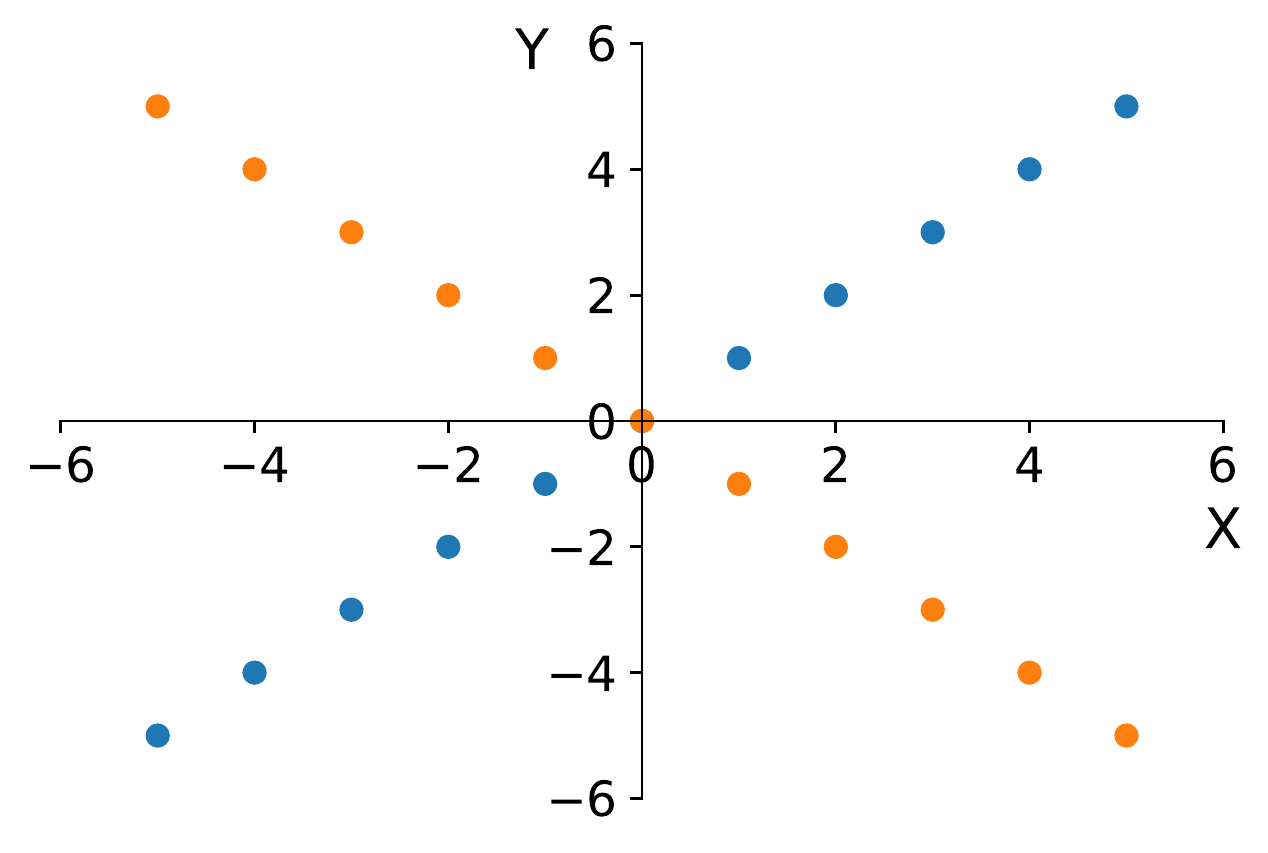}
        \caption{}
    \end{subfigure}
    \caption{(a) This plot shows an example of CLR applied to a simple dataset generated from $3$ different linear regressions. (b) This plot illustrates that it might be impossible to predict CLR labels from features $X$ only. In this case there are $2$ linear regressions that algorithm would find, depicted with orange and blue. However, ignoring $y$ will make points from different clusters indistinguishable. Since there is no information about cluster membership, the best possible prediction model in this case is a simple average of the two models, which is a line coinciding with x-axis.}
    \label{fig:pred-plot}
\end{figure}

The informal goal of CLR is to find clusters with different correlation patterns (simple illustrations of this are presented in Figure~\ref{fig:pred-plot}). Such clusters can be useful for the data analysis applications in a variety of different domains~\cite{bagirov2017prediction},~\cite{chirico2013clusterwise},~\cite{brusco2002simulated},~\cite{zhang2013explaining}. It seems natural that identifying such a relation would also be useful for prediction purposes. Indeed, if the correct cluster labels were known for the test points, CLR would provide a flexible non-linear and potentially discontinuous prediction function (Figure~\ref{fig:pred-plot}~(a)). The computational cost of fitting such a function would be similar to the usual linear regression by using, e.g. the standard approach proposed by Spath~\cite{spath1979algorithm} to solve problem~\ref{eq:clr}. However, for most datasets there is no easy way to assign CLR labels to the test points. It might even be impossible to do so, because CLR clusters are based on correlation information between features $X$ and targets $y$, which are not available for test points. A simple example showing an adversary situation for the restoration of CLR labels is depicted in Figure~\ref{fig:pred-plot} (b).

There are two existing approaches to overcome this problem. The first approach is to modify CLR objective, so that the function $l(x)$ predicting cluster labels for test points is always defined and is easy to compute. One way to do it was proposed by Manwani and Sastry~\cite{manwani2015k}. The authors suggest a k-plane regression algorithm, which combines CLR objective function with k-means~\cite{macqueen1967some} objective, thus requiring final clusters to become more separable. Leveraging this at test time, the unknown labels can be obtained by associating each point with the closest cluster center, as done in the usual k-means algorithm. It should be noted that this label assignment technique is a heuristic and may not restore the correct clusters even for the training points, since CLR and k-means objectives are often contradicting each other.

The second approach is to assign a certain weight $v_i(x)$ to each cluster $i$. In this case, the final prediction function is the weighted average of the predictions from all cluster models. 
For example, for the case presented in Figure~\ref{fig:pred-plot} (b), the best possible prediction model would be a line coinciding with the x-axis, which can be obtained with $v_1(x) = v_2(x) = 0.5$. Informally this indicates that there is no information to prefer one cluster versus the other for any point and thus a simple average of the two models is used as the final prediction. An incarnation of this approach with $v_i(x) = \sfrac{\abr{C_i}}{n}$\footnote{Where $\abr{C_i}$ denotes the number of elements in cluster $i$.} was used by Bagirov et al. for prediction of monthly rainfall in Australia~\cite{bagirov2017prediction}. This weighting strategy is also a heuristic and as we show in section~\ref{seq:weight} the final prediction model that uses it is confined to be linear.

In this paper we propose two alternative methods for both of the aforementioned approaches. The first method that we propose is to replace a heuristic-based label prediction function $l(x)$ or cluster weights functions $v_i(x)$ with the functions learned from the data. Namely, $l(x)$ can be any classification model which is trained on the objects $x_i$ and their corresponding CLR labels which are known for the training set. When the classification model can yield the probabilities for each cluster, these probabilities can be used as the weights $v_i(x)$. We call this family of models \textit{predictive CLR} (CLR-p). We experiment with a number of variations of CLR-p models, including adding k-means objective, using different CLR clustering algorithms, using different label prediction models and replacing linear regression inside CLR with Ridge regression or Lasso regression~\cite{tibshirani1996regression}. We also found that building ensembles of CLR-p models by starting CLR clustering from different random initializations can significantly improve prediction results.

The second method that we propose is to change CLR objective by incorporating some natural constraints often available in the data. For example, consider a dataset containing medical claims of patients. It is reasonable to assume that patients treated at the same hospital should have similar prediction models. This can be achieved by constraining CLR to put all patients from the same hospital into a single cluster. Since the hospital membership is available at test time there is no problem of obtaining test CLR labels. This results in a very simple change to the CLR algorithm which is enough to turn it into a powerful prediction model, which is slower than linear regression by only a constant (i.e. not depending on the dataset size) factor. We call this algorithm \textit{constrained CLR} (CLR-c).

We evaluate the two proposed algorithms on three public datasets from UCI ML repository~\cite{asuncion2007uci} as well as on the large corpus of medical claims of patients from a particular health insurance provider. Overall, we found that both CLR-c and CLR-p significantly improve the results over the standard linear regression, while being only $\approx 10$-$100$ times slower\footnote{This is achieved by running underlying CLR clustering for only $5$ iterations which turned out to be enough in most cases. However, it is still necessary to fit one linear regression to each cluster which increases time further.}. To achieve this run-time for CLR-p, it is necessary to use a fast classification model, e.g. logistic regression. However, we found that CLR-p can be made much more powerful by using random forest~\cite{breiman2001random} as the label prediction model and by using ensembling over different random initializations. This slows down the computation, but the algorithm still scales well to large datasets as we demonstrate in our experiments. We also show that both CLR-p and CLR-c outperform k-plane regression~\cite{manwani2015k} as well as the weighting approach based on the cluster sizes used by Bagirov et al.~\cite{bagirov2017prediction} on all four datasets examined in this paper. Finally, we compare the proposed methods to two popular regression algorithms: kernel support vector regression (SVR)~\cite{drucker1997support} and random forest. On the UCI ML datasets, CLR-p\footnote{With random forest as the label prediction model and ensembling over different random initializations.} algorithm shows comparable performance to support vector regression and outperforms random forest. On the health insurance data, CLR-c shows the best performance in terms of both, the final accuracy as well as the run-time of the method, by leveraging hospital-based constraints available in the data.

\section{Related work}
One of the first attempts to use CLR for prediction dates back to the work of Kang et al. which suggests to use fuzzy clustering with Dirichlet prior on cluster membership~\cite{kang2009clusterwise}. Making the prior dependent on features $X$ only, the authors obtain a natural way to derive CLR labels at test time. Although this approach is theoretically appealing, the algorithm involves expensive Gibbs sampling procedure which limits its application to relatively small datasets. The authors do not explore any application to real-world datasets providing the experiments only on synthetically generated data.

Manwani and Sastry~\cite{manwani2015k} propose k-plane regression method that combines CLR with k-means and uses the closest cluster centers as labels at test time. They compare their method to SVR and to hinging hyperplane method~\cite{breiman1993hinging} on synthetic data as well as on the UCI ML datasets. The authors show that their method is much faster than SVR; however it produces significantly worse prediction results. 

Bagirov et al. applies a modification of CLR~\cite{bagirov2015algorithm} to prediction of monthly rainfall in Australia, taking weighted average of different models based on the cluster sizes~\cite{bagirov2017prediction}. However, since the cluster weights do not depend on the features of the new object, the final model is always linear which limits its representational power.

The idea of performing constrained CLR was first proposed in~\cite{plaia2005constrained}; however, the authors do not explore potential applications to prediction.

\section{CLR clustering algorithms}
In this paper we experiment with two basic algorithms for finding CLR clusters on the training data.
\subsection{Hard CLR}
The first algorithm for finding CLR clusters is a k-means regularized version of the original CLR which was first introduced by Brusco et al.~\cite{brusco2003multicriterion} and later rediscovered and applied for prediction by Manwani and Sastry~\cite{manwani2015k}. The algorithm starts with the random assignment of cluster labels and sequentially iterates two steps: regression and labeling. On the regression step, given fixed labels assignment, the linear regression problem is solved for each cluster separately:
\begin{equation}\label{eq:hard1}
    \text{For each cluster $C_i$ solve:}\ \min_{w_i, b_i}\sum_{j \in C_i}\rbr{y_j - x_j^Tw_i - b_i}^2.
\end{equation}
On the labeling step, given fixed regression coefficients each object is reassigned to the best cluster:
\begin{equation}\label{eq:hard2}
    \text{Assign $\{x_j, y_j\}$ to cluster}\ c = \argmin_{i}\sbr{\rbr{y_j - x_j^Tw_i - b_i}^2 + \gamma\nbr{ x_j - \frac{1}{\abr{C_i}}\sum_{t \in C_i}x_t}_2^2}.
\end{equation}
The first term in equation~\ref{eq:hard2} is the usual CLR objective and the second term is k-means regularization. If the regularization parameter $\gamma$ is set to zero, the algorithm becomes equivalent to the vanilla CLR. If $\gamma \to \infty$, the algorithm is equivalent to k-means. We will refer to this algorithm as \textit{hard CLR} through the rest of the paper.

Additionally, hard CLR can be straightforwardly extended to incorporate different regression functions. For example, adding $L_2$ or $L_1$ regularization to the minimization problem on regression step, would be equivalent to switching linear regression with Ridge regression or Lasso regression. More generally, replacing $x_j^Tw_i + b_i$ with $f(x_j)$ in both regression and labeling steps allows the algorithm to use any regression function $f(x)$.

Finally, hard CLR algorithm can be easily modified to accommodate any user specified constraints. Formally, the definition of the constraints is the same as definition of the clusters: the constraints set consists of $m$ disjoint sets $S_i$, covering all data points. Then, in order to preserve the invariant that all points from one constraint set have to go to the same cluster, the labeling step of hard CLR needs to be modified in the following way:

\begin{align*}
    &\text{Assign all objects $\cbr{x_j, y_j} \in S_q$ to cluster}\ c \\ &c = \argmin_{i}\sum_{j \in S_q}\sbr{\rbr{y_j - x_j^Tw_i - b_i}^2 + \gamma\nbr{ x_j - \frac{1}{\abr{C_i}}\sum_{t \in C_i}x_t}_2^2}.
\end{align*}

\subsection{Soft CLR}
The second approach which we use to obtain CLR clusters on training data is to perform the fuzzy CLR proposed by DeSarbo et al.~\cite{desarbo1988maximum}. The algorithm assumes that the conditional distribution of $y$ given $x$ is a mixture of Gaussian distributions that can be written in the following way

\begin{equation}\label{eq:softCLR}
    p(y|x) = \sum_{i=1}^k\pi_i\Nm\rbr{x^Tw_i + b_i, \sigma_i^2}.
\end{equation}
The unknown parameters $\pi_i, w_i$ and $\sigma_i$ can be estimated using EM-algorithm. In this case, k-means regularization term can be viewed as assuming a Gaussian prior on $p(x) = \sum_{i=1}^k\pi_i\Nm(\mu_i, \frac{\sigma_i^2}{\gamma})$. Exact EM equations are presented in the Appendix~\ref{app:em}. Note that in this formulation CLR finds soft assignment of clusters which are computed on E-step. Even though this soft assignment contains more information that could potentially be useful for prediction, in this work we will always convert it to hard assignment by associating each point with the cluster that has maximum probability. We will call this algorithm \textit{soft CLR} through the rest of the paper.

\section{Prediction methods}
Regardless of the choice of CLR algorithm, there are two basic options how to use the found clusters for prediction. 
\subsection{Predicting test labels}
The first option is to derive the cluster labels for unseen objects using some function $l(x)$ and then use the corresponding linear regression models for prediction. For example in the case of k-plane regression 

\[ l(x) = \argmin_i\nbr{x - \frac{1}{\abr{C_i}}\sum_{t \in C_i}x_t}_2^2. \]

For the CLR-p method we propose to learn the function $l(x)$ from the data. In this paper we experiment with logistic regression and random forest as the label prediction models.

For the CLR-c method $l(x)$ is a simple look-up function that returns cluster label of the constraint set $S_i$ to which $x$ belongs to.
\subsection{Weighting all models}\label{seq:weight}
The second way to obtain predictions for unseen test objects is by assigning a certain weight $v_i(x)$ to each cluster $i$. After that the prediction is obtained by averaging the results from all cluster models:
\begin{equation}\label{eq:wpred}
    f(x) = \sum_{i=1}^kv_i(x)\rbr{x^Tw_i + b_i}.
\end{equation}

When $v_i(x) = \pi_i$ we recover the EM approximation of equation~\ref{eq:softCLR}:
\begin{equation*}
    \max_y p(y|x) \approx f(x) = \sum_{i=1}^k\pi_i\rbr{x^Tw_i + b_i} = x^T\sum_{i=1}^k\pi_iw_i + \sum_{i=1}^k\pi_ib_i = x^T\widehat{w} + \widehat{b},
\end{equation*}
which can be used for prediction, although as shown above the final prediction will still be linear. The same holds for $v_i(x) = \frac{\abr{C_i}}{n}$ when the prediction rule~\ref{eq:wpred} becomes equivalent to the approach used by Bagirov et al.~\cite{bagirov2017prediction}. In general, $f(x)$ will be linear for any $v_i(x)$ that is fixed for all points in a cluster, i.e. it does not depend on $x$. 

For the CLR-p algorithm the weights $v_i(x)$ can be naturally obtained as probabilities that model $l(x)$ assigns to each cluster. The k-plane regression algorithm can be straightforwardly modified to use weighting approach. For that we propose to use normalized distances to cluster centers as weights. That is
\[ v_i(x) = \frac{\nbr{x - \frac{1}{\abr{C_i}}\sum_{t \in C_i}x_t}_2}{\sum_{j=1}^k\nbr{x - \frac{1}{\abr{C_j}}\sum_{t \in C_j}x_t}_2}. \]

\subsection{Ensembling}
We noticed that running CLR with different random initializations would often produce drastically different cluster assignments, which leads to a high variance of the prediction results. To lower this variance and improve the final accuracy we propose to build an ensemble of CLR-based models, averaging the prediction results across different random initializations.

\section{Datasets}
To evaluate all the proposed techniques we used 3 datasets from UCI ML repository: Boston Housing, Abalone and Auto-mpg. We use the same datasets as used in the k-plane regression paper~\cite{manwani2015k} except the Computer Activity dataset which we were not able to find online. We also tried to follow the same preprocessing, which was not always possible since some important details are omitted from the paper. It should be noted that since there are no natural constraints available in this case, we simply used the most suited categorical feature for each dataset (or created one, as in the case of Abalone) in order to evaluate CLR-c approach. The only criteria that we used to choose which features to constraint on was that they should have not too big and not too small number of distinct values to provide enough flexibility for the final prediction model. For the Auto-mpg dataset we use ``model year'', for the Boston Housing we use ``index of accessibility to radial highways'' and for the Abalone dataset we use an additionally added binning of the ``diameter'' feature as the constraints. More detailed description of these datasets can be found online and the exact preprocessing steps that we take are described in the Appendix~\ref{app:datasets}. 

In addition to the relatively small UCI ML datasets, we evaluate all of the proposed techniques on the large health insurance dataset. This dataset consists of medical claims of patients from a particular health insurance provider. After some preprocessing we obtain $\approx 400000$ claims each characterized with $146$ features. This dataset is quite big, but at the same time very sparse, with $94\%$ of all feature values being zeros. We use hospital membership as the constraints for this data. More details about preprocessing and description of the features is available in the Appendix~\ref{app:health}.

\section{Experiments}
The Python implementation of all experiments described in this paper is available online\footnote{\href{https://github.com/Kipok/clr_prediction}{\nolinkurl{github.com/Kipok/clr_prediction}}}. All algorithms are implemented following scikit-learn base estimator API~\cite{buitinck2013api}. 

\subsection{CLR-p vs k-plane regression}\label{clrpvskplane}

For the first experiment we decided to compare k-plane regression model with CLR-p in the controlled environment when the CLR labels are known for the test points. For that we took the health insurance dataset, split it into train ($75\%$) and test ($25\%$) sets, but ran hard version of CLR on the whole data together (for $20$ iterations with different number of clusters and different values of k-means regularization parameter $\gamma$). Since CLR was run on the whole data, we had an access to CLR labels for test points and thus could estimate how well all models can predict those labels and how well they perform overall. It should be noted that the only potential overfitting in this setup comes from the fact that we run CLR on the whole data. But after that we still ``forget'' the found labels and evaluate all models fairly without looking at the test targets or the found CLR labels (except to evaluate the accuracy of labels prediction). Table~\ref{tab:clrp_kplane} contains the complete results of this experiment. The second column shows the coefficient of determination ($R^2$) of the ``best'' model for which we used the correct CLR labels. The next two columns show the label prediction accuracy for CLR-p (with random forest) and k-plane regression correspondingly. The final four columns show the test $R^2$ for the non-weighted and weighted CLR-p and k-plane regression methods.

\begin{table}[t]
  \caption{Comparison of k-plane regression and CLR-p. The baseline $R^2$ of linear regression is $0.3$.}
  \label{tab:clrp_kplane}
  \centering
  {\renewcommand{\arraystretch}{1.5}
  \begin{tabular}[t]{|c|c|c|c|c|c|c|c|}
  \hline  
  k, $\gamma$ & $R^2$ ``best'' & \makecell{$\text{CLR}_p$ label \\ accuracy} & \makecell{k-plane label \\ accuracy} & \makecell{$\text{CLR}_p$ \\ $R^2$} & \makecell{k-plane \\ $R^2$} & \makecell{weighted \\ $\text{CLR}_p$ $R^2$} & \makecell{weighted \\ k-plane $R^2$} \\
  \hline  
    2, 0 & 0.77 & 0.87 & 0.80 & 0.21 & -0.15 & {\bf 0.40} & -0.22 \\
  \hline  
    2, 10 & 0.76 & 0.91 & 0.81 & 0.20 & -0.56 & 0.39 & -0.82 \\
  \hline  
    2, 50 & 0.54 & 0.95 & 0.93 & 0.24 & 0.05 & 0.34 & -0.44 \\
  \hline  
    2, 100 & 0.55 & 0.97 & 0.97 & 0.30 & 0.28 & 0.36 & -2.04 \\
  \hline  
    2, 1000 & 0.33 & 0.99 & 1.00 & 0.33 & 0.33 & 0.33 & 0.24 \\
  \hline 
  \hline  
    8, 0 & 0.96 & 0.38 & 0.26 & 0.09 & -0.17 & 0.38 & -0.24 \\
  \hline  
    8, 10 & 0.88 & 0.86 & 0.86 & 0.19 & 0.13 & {\bf 0.39} & -1.50 \\
  \hline  
    8, 50 & 0.60 & 0.97 & 0.97 & 0.30 & 0.28 & 0.36 & -1.63 \\
  \hline  
    8, 100 & 0.53 & 0.98 & 0.98 & 0.32 & 0.30 & 0.35 & -0.22 \\
  \hline  
    8, 1000 & 0.34 & 0.99 & 1.00 & 0.33 & 0.33 & 0.34 & 0.24 \\
  \hline  
  \end{tabular}}
\end{table}

We can draw a few important observations from this experiment. First, when the true CLR labels are known, the potential CLR-based prediction accuracy is unusually high for the small values of $\gamma$. This is not surprising since it was observed by Brusco et al.~\cite{brusco2008cautionary} that CLR can find clusters with very high values of $R^2$ even when $X$ and $y$ are generated independently (by grouping points with similar targets together). Thus, these $R^2$ values provide a very loose upper-bound on the performance of the CLR-based prediction methods, since it is usually not possible to find the correct labels for all points without accessing test targets. By looking at the next two columns we can see that as $\gamma$ increases, this upper-bound goes down, while label prediction accuracy of both methods goes up. This is expected since k-means regularization makes clusters more separable (i.e. easier to predict cluster labels), but at the same time it counteracts CLR objective which results in worse linear regression fit inside each cluster. Interestingly, even though label prediction accuracy is quite high for both methods (more than $80\%$ in most cases), the final prediction models are not good when non-weighted approach is used. This shows that using an incorrect cluster model can lead to predictions that are much worse than using one joint model for all clusters or even than using a simple average of all targets (which results in a negative $R^2$ for some of the k-plane regression models). For CLR-p this can be fixed by using weighted approach. We expected the same pattern for k-plane regression, but the current weighting strategy does not work well. Finally, we can notice that CLR-p has higher label prediction accuracy when $\gamma$ is relatively small which results in significantly better final prediction models. This provides a clear evidence that learning the label prediction function from the data is a better strategy than using a heuristic label prediction approach of the k-plane regression method.

\begin{figure}[t]
    \centering
    \begin{subfigure}{0.49\textwidth}
        \centering
        \includegraphics[width=1.0\textwidth]{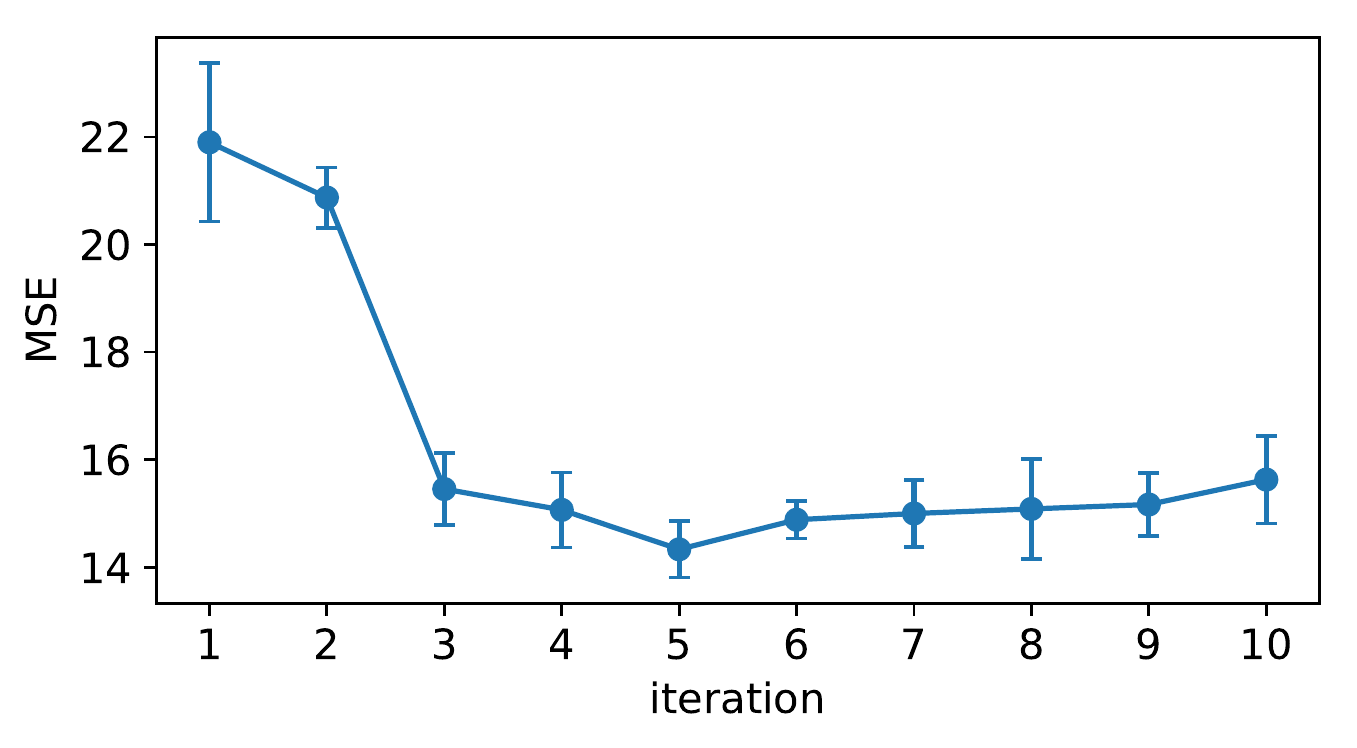}
        \caption{Dependence on number of CLR iterations}
    \end{subfigure}
    \begin{subfigure}{0.49\textwidth}
        \centering
        \includegraphics[width=1.0\textwidth]{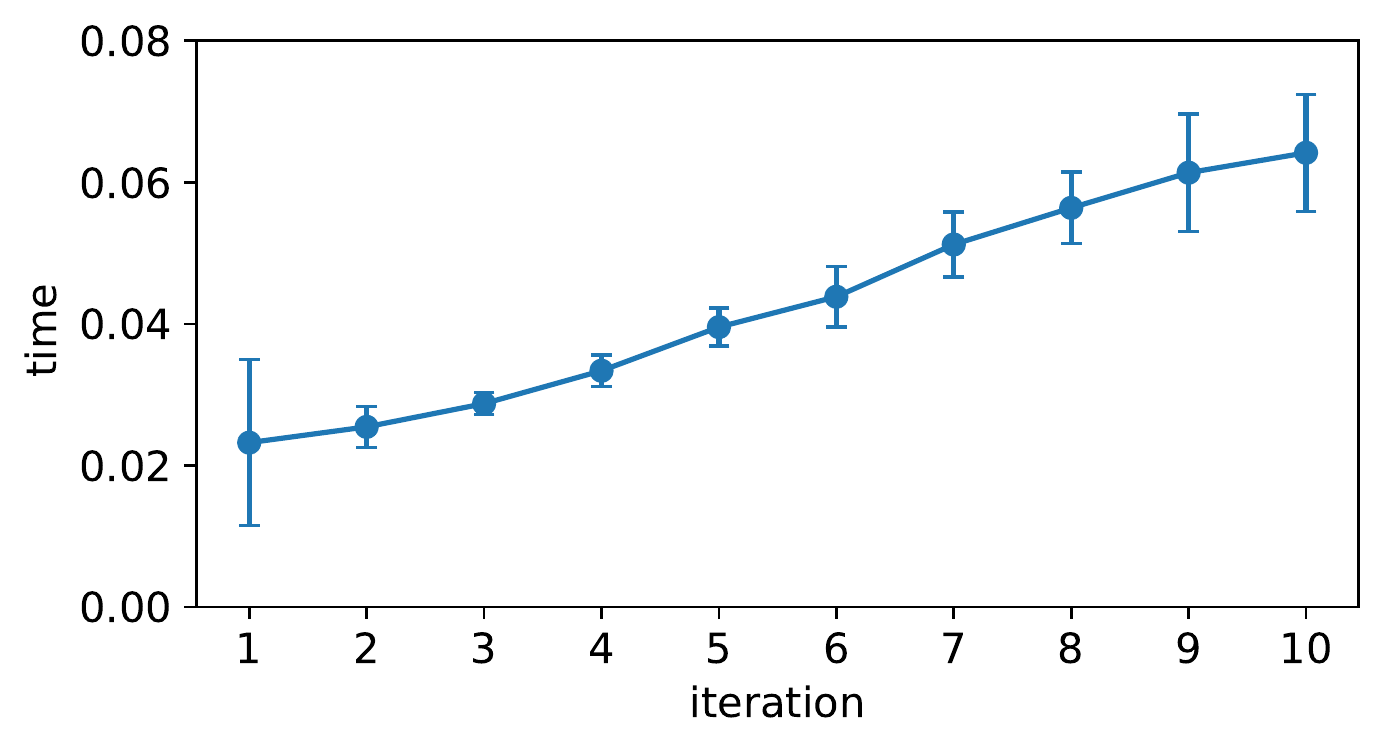}
        \caption{Dependence on number of CLR iterations}
    \end{subfigure}
    \begin{subfigure}{0.49\textwidth}
        \centering
        \includegraphics[width=1.0\textwidth]{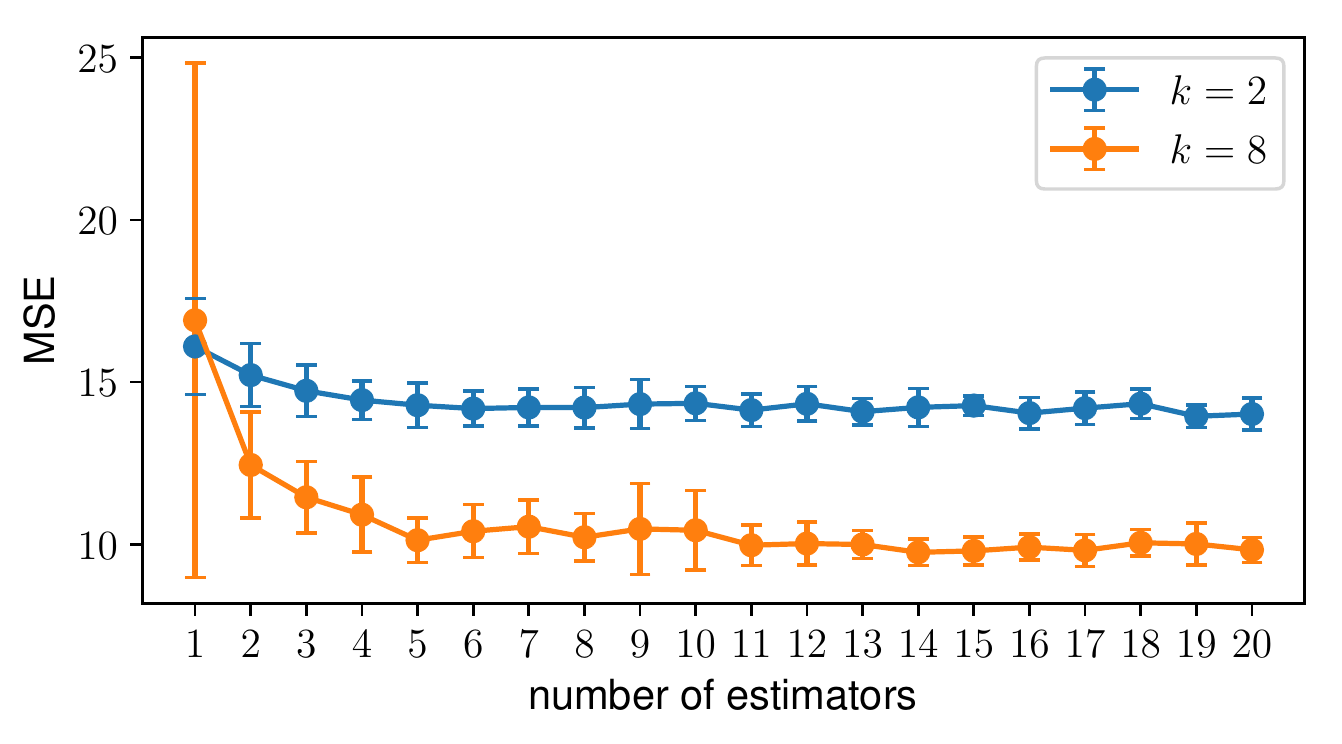}
        \caption{Dependence on ensemble size (CLR-p)}
    \end{subfigure}
    \begin{subfigure}{0.49\textwidth}
        \centering
        \includegraphics[width=1.0\textwidth]{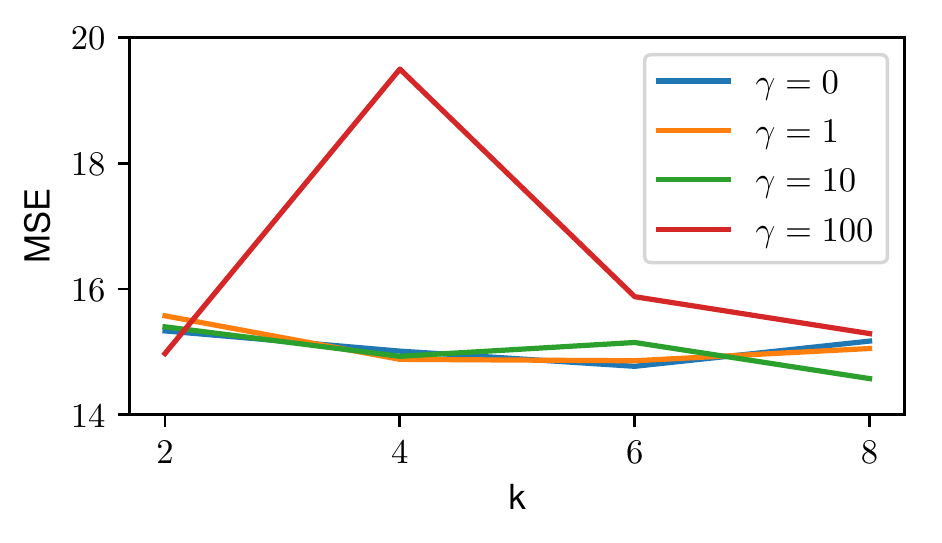}
        \caption{Performance of CLR-c method}
    \end{subfigure}
    \caption{(a), (b) shows dependence of MSE and time on the number of CLR iterations for CLR-p method on Boston dataset ($k=8$ clusters with $\gamma = 10$) . (c) shows dependence of MSE on the ensemble size of CLR-p method on Boston dataset ($\gamma=10$). (d) shows dependence of the final MSE on the different parameters for CLR-c method.}
    \label{fig:more-params}
\end{figure}

\begin{figure}[t]
    \centering
    \includegraphics[width=1.0\textwidth]{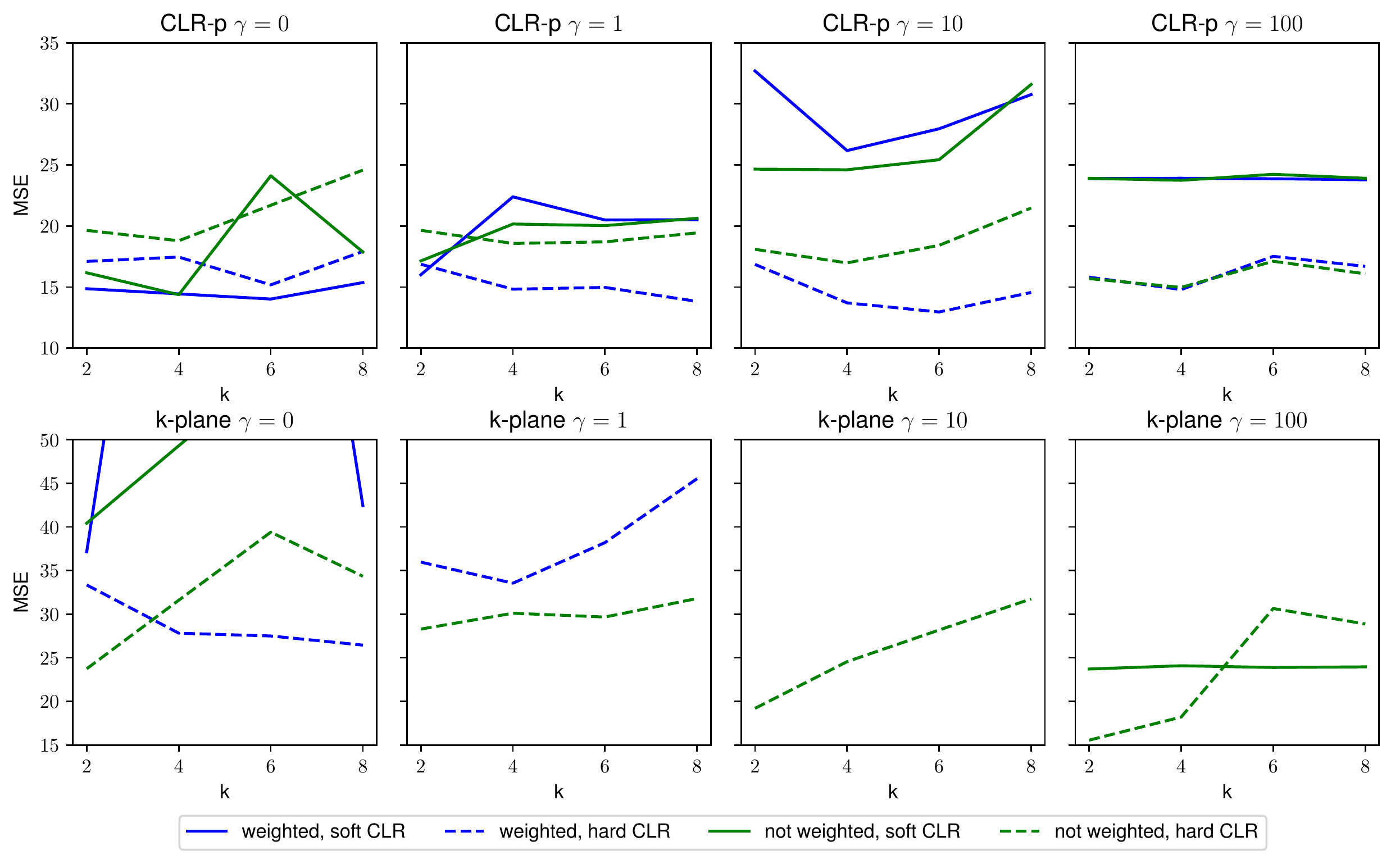}
    \caption{Boston housing dataset, dependence on parameters for CLR-p and k-plane regression. $k$ is the number of clusters.}
    \label{fig:params-boston}
\end{figure}

\subsection{Dependence on parameters}
In the next series of experiments we explored how different parameters affect the quality of the final prediction models for different methods. We present the results on Boston dataset only, because of the space constraints, although the main trends are the same for all datasets explored in this paper (complete graphs are available in the Appendix~\ref{app:dep}). For every parameter combination we ran 5 repetitions of 10-fold cross-validation. We report mean values as well as the standard deviations which are calculated only over the repetitions of cross-validation (meaning that we first aggregate results from 10 folds and then only compute the variance over the $5$ numbers). The non-aggregated standard deviation is significantly higher for many models, which is likely due to the small dataset sizes. The aggregated numbers provide better comparison of the models with each other as well as with the results reported by Manwani and Sastry~\cite{manwani2015k}.

Figure~\ref{fig:more-params} (a),(b),(c) shows the dependence of CLR-p method on the number of iterations and on the ensemble size. We can see that about $5$ iterations is enough to reach peak Mean Squared Error (MSE) (this holds for a variety of different parameters, not only for the presented case). Increasing ensemble size helps to decrease both average MSE and it's variance. Figure~\ref{fig:more-params} (d) and Figure~\ref{fig:params-boston} shows the dependence of all methods on the remaining parameters. We can see that CLR-c model shows almost the same accuracy for all variations of the parameters. For CLR-p the best parameters are weighted, hard CLR with medium $\gamma = 1$ or $10$. For k-plane regression the best parameters are not-weighted, hard CLR with as high $\gamma$ as possible and small number of clusters ($2$~or~$4$). The fact that k-plane regression works best with as high $\gamma$ as possible means that most of the gain comes from the use of k-means algorithm and not from CLR. Thus, it is likely that fitting a separate linear regression to the usual k-means clusters would be as effective as using k-plane regression. As can be seen from our experiments, the same is not true for CLR-p.

\subsection{Evaluation of the best models}
For the final experiment we ran an extensive grid search on all four datasets for the CLR-p, CLR-c, k-plane regression, support vector regression (SVR), random forest (RF) and linear regression. Each parameters configuration was evaluated using 5 repetitions of 10 fold cross-validation. To make sure that no overfitting occurs we repeated the entire grid-search with different random seed and the results did not change significantly. The exact parameters of the grid search are presented in the Appendix~\ref{app:expsetup}. Table~\ref{tab:eval} contains the results of this experiment (RF stands for random forest, LR stands for logistic regression and ``ens'' means that ensemble of $10$ models was used). We can see that on UCI ML datasets, CLR-p performs on par with SVR\footnote{It shows better performance on $2$ out of $3$ datasets, however the difference in results is smaller than one standard deviation.}. It outperforms random forest by a fair margin on Boston housing and Auto-mpg and outperforms k-plane and linear regression by a big margin on all datasets. However, this comes at a cost of a high run-time of the method\footnote{Which can be solved to some extent by running different models in parallel for the ensembles. This will roughly make the run-time of ensembles the same as the run-time of the base models which is also presented in the Table~\ref{tab:eval}.}. Fast logistic regression version of CLR-p does not perform well enough which shows that high quality label prediction is one of the key ingredients for the success of CLR-based methods. CLR-c method also performs worse than SVR or RF, however shows sizable improvement over k-plane regression on $2$ out of $3$ datasets and is very fast to run.

For the medical claims datasets CLR-c approach yields the best performance, while CLR-p performs worse than random forest, but still significantly better than the k-plane regression approach. Such a good performance of CLR-c algorithm is likely justified by the good quality of the constraint set (hospital membership), since it is expected that the claims inside single hospital would have similar regression models. The number of hospitals is also large enough ($\approx 2000$) to provide reasonable flexibility of the final prediction model. Notably, the best performance of k-plane regression algorithm is achieved with $\gamma = 10^5$, in which case the underlying CLR is almost identical to k-means algorithm. Any smaller value of $\gamma \in \cbr{0, 1, 10, 100, 1000}$ resulted in a significant decrease in prediction accuracy, making k-plane regression perform worse than linear regression. To achieve the presented results, it was also necessary to run k-plane regression training for $20$ iterations (instead of $5$) which also made it $\approx 4$ times slower.

\begin{table}[t]
  \caption{Evaluation of the proposed algorithms}
  \label{tab:eval}
  \centering
  {\renewcommand{\arraystretch}{1.2}
  \begin{tabular}[t]{|c|c|c|c|}
  \multicolumn{4}{c}{Abalone} \\
  \hline                   & MSE              & $R^2$  & time  \\
  \hline  CLR-p RF ens     & $4.53 \pm 0.03$  & $0.56$ & $3.28$ \\
  \hline  RF               & $4.56 \pm 0.01$  & $0.56$ & $1.56$ \\
  \hline  SVR              & $4.56 \pm 0.01$  & $0.56$ & $1.67$ \\ 
  \hline  CLR-c ens        & $4.59 \pm 0.01$  & $0.56$ & $0.82$ \\
  \hline  CLR-c            & $4.63 \pm 0.03$  & $0.55$ & $0.09$ \\
  \hline  k-plane ens      & $4.67 \pm 0.03$  & $0.55$ & $0.86$ \\ 
  \hline  k-plane          & $4.77 \pm 0.03$  & $0.54$ & $0.09$ \\ 
  \hline  CLR-p LR ens     & $4.78 \pm 0.18$  & $0.54$ & $2.40$ \\
  \hline  CLR-p RF         & $4.80 \pm 0.01$  & $0.54$ & $0.33$ \\
  \hline  CLR-p LR         & $4.82 \pm 0.20$  & $0.53$ & $0.19$ \\
  \hline  LR               & $4.98 \pm 0.01$  & $0.52$ & $0.004$ \\
  \hline
  \end{tabular}}
  {\renewcommand{\arraystretch}{1.2}
  \begin{tabular}[t]{|c|c|c|c|}
  \multicolumn{4}{c}{Boston housing} \\
  \hline                   & MSE             & $R^2$  & time \\
  \hline  CLR-p RF ens     & $9.3 \pm 0.5$   & $0.89$ & $1.40$  \\
  \hline  SVR              & $9.6 \pm 0.3$   & $0.88$ & $0.14$  \\
  \hline  RF               & $10.3 \pm 0.6$  & $0.87$ & $0.50$  \\
  \hline  CLR-c ens        & $13.5 \pm 0.3$  & $0.84$ & $0.18$  \\
  \hline  CLR-p RF         & $13.5 \pm 1.1$  & $0.83$ & $0.15$  \\
  \hline  CLR-p LR ens     & $13.9 \pm 0.4$  & $0.83$ & $0.30$  \\
  \hline  k-plane ens      & $13.9 \pm 0.8$  & $0.83$ & $0.36$  \\
  \hline  CLR-c            & $15.0 \pm 0.5$  & $0.81$ & $0.03$  \\
  \hline  CLR-p LR         & $15.3 \pm 0.4$  & $0.81$ & $0.01$  \\
  \hline  k-plane          & $15.8 \pm 0.5$  & $0.80$ & $0.01$  \\
  \hline  LR               & $23.9 \pm 0.2$  & $0.70$ & $0.001$ \\
  \hline
  \end{tabular}}
  {\renewcommand{\arraystretch}{1.2}
  \begin{tabular}[t]{|c|c|c|c|}
  \multicolumn{4}{c}{Auto-mpg} \\
  \hline                   & MSE              & $R^2$  & time  \\
  \hline  SVR              & $6.55 \pm 0.08$  & $0.89$ & $0.02$ \\
  \hline  CLR-p RF ens     & $6.76 \pm 0.10$  & $0.89$ & $1.37$ \\ 
  \hline  CLR-p RF         & $7.47 \pm 0.33$  & $0.87$ & $0.13$ \\
  \hline  RF               & $7.53 \pm 0.16$  & $0.87$ & $0.47$ \\
  \hline  k-plane ens      & $7.81 \pm 0.17$  & $0.87$ & $0.42$ \\ 
  \hline  CLR-p LR ens     & $8.17 \pm 0.32$  & $0.86$ & $0.52$ \\
  \hline  CLR-c ens        & $8.45 \pm 0.11$  & $0.86$ & $0.21$ \\
  \hline  CLR-p LR         & $8.87 \pm 0.39$  & $0.85$ & $0.03$ \\
  \hline  CLR-c            & $9.41 \pm 0.35$  & $0.84$ & $0.01$ \\
  \hline  k-plane          & $9.64 \pm 0.65$  & $0.84$ & $0.01$ \\ 
  \hline  LR               & $11.34 \pm 0.09$  & $0.81$ & $0.001$ \\
  \hline
  \end{tabular}}
  {\renewcommand{\arraystretch}{1.2}
  \begin{tabular}[t]{|c|c|c|c|}
  \multicolumn{4}{c}{Medical claims} \\
  \hline                   & MSE            & $R^2$  & time    \\
  \hline  CLR-c ens        & $39.0 \pm 0.1$ & $0.47$ & $31$    \\ 
  \hline  CLR-c            & $40.0 \pm 0.3$ & $0.45$ & $24$    \\ 
  \hline  RF               & $43.0 \pm 0.1$ & $0.41$ & $272$   \\ 
  \hline  CLR-p RF ens     & $45.3 \pm 0.5$ & $0.38$ & $3340$  \\ 
  \hline  CLR-p RF         & $45.4 \pm 0.8$ & $0.38$ & $465$  \\ 
  \hline  k-plane ens      & $47.6 \pm 0.5$ & $0.35$ & $115$   \\ 
  \hline  CLR-p LR ens     & $48.2 \pm 0.3$ & $0.34$ & $359$  \\ 
  \hline  k-plane          & $48.6 \pm 0.5$ & $0.33$ & $109$   \\ 
  \hline  CLR-p LR         & $48.7 \pm 0.4$ & $0.33$ & $45$   \\ 
  \hline  LR               & $51.3 \pm 0.0$ & $0.30$ & $1$     \\
  \hline  SVR              & \multicolumn{3}{c|}{Did not finish in $15$ hours} \\
  \hline
  \end{tabular}}
\end{table}

\section{Discussion and conclusions}
Summarizing results of all experiments, we can see that the proposed algorithms compose a very flexible family of regression models. CLR-c algorithm is the most effective when there is a reasonable set of constraints available in the data, as was the case for the medical claims dataset. While additional research is required to explore how the constraints should be defined to increase predictive power of the method, we have shown that CLR-c can work with almost any kind of constraints. Even when the constraints are chosen almost arbitrary (as was the case with the UCI ML datasets) it can still show a significant improvement over linear regression, while being only slower by a constant (in the dataset size) factor, which is roughly equal to the number of clusters $\times$ number of CLR iterations. Thus, it can be used as an alternative to linear regression when the computational budget available to solve the problem is very restrictive. When the problem requires highly accurate predictions, ensemble of CLR-p models with random forest can be used instead. It often outperforms standard prediction methods, while being as scalable as the underlying classification model. Since in this paper we only explored logistic regression and random forest as the label prediction models, it might be possible that some other method will work even better (e.g. providing the same prediction accuracy, while being significantly faster). 

The key ingredient to the success of the CLR-p method is that we replace the heuristic label prediction approach used in k-plane regression with more principled idea of learning the label prediction function from the data. As shown in section~\ref{clrpvskplane} this alone significantly improves the final regression model. Using the learned label probabilities as cluster weights makes the results even better for CLR-p method, while using weighting approach for the k-plane regression algorithm usually makes the model much worse. This is likely because the weighting strategy for the k-plane regression method is based on the same distance-to-cluster-centers heuristic which is not well justified.

It should also be noted that both CLR-p and CLR-c do not require careful tuning and perform well with a wide range of different parameters. For example, for CLR-p algorithm we did not tune the underlying random forest on the UCI ML datasets\footnote{For the medical claims we did tuning for one set of parameters to make the model faster and used the found configuration for all other parameters in the grid search.}. Finally, we would like to note that both of the proposed methods (especially, CLR-c) might be as interpretable as linear regression, since the only difference is the prior CLR-based clustering, which usually has a very intuitive meaning. We did not explore interpretability of the methods in this paper, but it is a promising direction of the future research.

\bibliography{main}

\begin{thebibliography}{10}

\bibitem{asuncion2007uci}
A.~Asuncion.
\newblock Uci machine learning repository. university of california, irvine,
  school of information and computer sciences.
\newblock {\em http://mlearn. ics. uci. edu/MLRepository. html}, 2007.

\bibitem{bagirov2017prediction}
A.~M. Bagirov, A.~Mahmood, and A.~Barton.
\newblock Prediction of monthly rainfall in victoria, australia: Clusterwise
  linear regression approach.
\newblock {\em Atmospheric Research}, 188:20--29, 2017.

\bibitem{bagirov2015algorithm}
A.~M. Bagirov, J.~Ugon, and H.~G. Mirzayeva.
\newblock An algorithm for clusterwise linear regression based on smoothing
  techniques.
\newblock {\em Optimization Letters}, 9(2):375--390, 2015.

\bibitem{breiman1993hinging}
L.~Breiman.
\newblock Hinging hyperplanes for regression, classification, and function
  approximation.
\newblock {\em IEEE Transactions on Information Theory}, 39(3):999--1013, 1993.

\bibitem{breiman2001random}
L.~Breiman.
\newblock Random forests.
\newblock {\em Machine learning}, 45(1):5--32, 2001.

\bibitem{brusco2002simulated}
M.~J. Brusco, J.~D. Cradit, and S.~Stahl.
\newblock A simulated annealing heuristic for a bicriterion partitioning
  problem in market segmentation.
\newblock {\em Journal of Marketing Research}, 39(1):99--109, 2002.

\bibitem{brusco2008cautionary}
M.~J. Brusco, J.~D. Cradit, D.~Steinley, and G.~L. Fox.
\newblock Cautionary remarks on the use of clusterwise regression.
\newblock {\em Multivariate Behavioral Research}, 43(1):29--49, 2008.

\bibitem{brusco2003multicriterion}
M.~J. Brusco, J.~D. Cradit, and A.~Tashchian.
\newblock Multicriterion clusterwise regression for joint segmentation
  settings: An application to customer value.
\newblock {\em Journal of Marketing Research}, 40(2):225--234, 2003.

\bibitem{buitinck2013api}
L.~Buitinck, G.~Louppe, M.~Blondel, F.~Pedregosa, A.~Mueller, O.~Grisel,
  V.~Niculae, P.~Prettenhofer, A.~Gramfort, J.~Grobler, et~al.
\newblock Api design for machine learning software: experiences from the
  scikit-learn project.
\newblock {\em arXiv preprint arXiv:1309.0238}, 2013.

\bibitem{chirico2013clusterwise}
P.~Chirico.
\newblock A clusterwise regression method for the prediction of the disposal
  income in municipalities.
\newblock In {\em Classification and Data Mining}, pages 173--180. Springer,
  2013.

\bibitem{desarbo1988maximum}
W.~S. DeSarbo and W.~L. Cron.
\newblock A maximum likelihood methodology for clusterwise linear regression.
\newblock {\em Journal of classification}, 5(2):249--282, 1988.

\bibitem{desarbo1989simulated}
W.~S. DeSarbo, R.~L. Oliver, and A.~Rangaswamy.
\newblock A simulated annealing methodology for clusterwise linear regression.
\newblock {\em Psychometrika}, 54(4):707--736, 1989.

\bibitem{drucker1997support}
H.~Drucker, C.~J. Burges, L.~Kaufman, A.~J. Smola, and V.~Vapnik.
\newblock Support vector regression machines.
\newblock In {\em Advances in neural information processing systems}, pages
  155--161, 1997.

\bibitem{fetter1986diagnosis}
R.~B. Fetter and J.~L. Freeman.
\newblock Diagnosis related groups: product line management within hospitals.
\newblock {\em Academy of Management Review}, 11(1):41--54, 1986.

\bibitem{kang2009clusterwise}
C.~Kang and S.~Ghosal.
\newblock Clusterwise regression using dirichlet mixtures.
\newblock {\em Advances in multivariate statistical methods}, 4:305, 2009.

\bibitem{macqueen1967some}
J.~MacQueen et~al.
\newblock Some methods for classification and analysis of multivariate
  observations.
\newblock In {\em Proceedings of the fifth Berkeley symposium on mathematical
  statistics and probability}, volume~1, pages 281--297. Oakland, CA, USA.,
  1967.

\bibitem{manwani2015k}
N.~Manwani and P.~Sastry.
\newblock K-plane regression.
\newblock {\em Information Sciences}, 292:39--56, 2015.

\bibitem{megiddo1982complexity}
N.~Megiddo and A.~Tamir.
\newblock On the complexity of locating linear facilities in the plane.
\newblock {\em Operations research letters}, 1(5):194--197, 1982.

\bibitem{paige1982lsqr}
C.~C. Paige and M.~A. Saunders.
\newblock Lsqr: An algorithm for sparse linear equations and sparse least
  squares.
\newblock {\em ACM transactions on mathematical software}, 8(1):43--71, 1982.

\bibitem{plaia2005constrained}
A.~Plaia.
\newblock Constrained clusterwise linear regression.
\newblock {\em New Developments in Classification and Data Analysis}, pages
  79--86, 2005.

\bibitem{spath1979algorithm}
H.~Sp{\"a}th.
\newblock Algorithm 39 clusterwise linear regression.
\newblock {\em Computing}, 22(4):367--373, 1979.

\bibitem{tibshirani1996regression}
R.~Tibshirani.
\newblock Regression shrinkage and selection via the lasso.
\newblock {\em Journal of the Royal Statistical Society. Series B
  (Methodological)}, pages 267--288, 1996.

\bibitem{zhang2013explaining}
W.~Zhang and P.~L. Durango-Cohen.
\newblock Explaining heterogeneity in pavement deterioration: Clusterwise
  linear regression model.
\newblock {\em Journal of Infrastructure Systems}, 20(2):04014005, 2013.

\end{thebibliography}
\bibliographystyle{abbrv}

\newpage
\appendix
\section*{Appendix}

\section{Datasets}\label{app:datasets}
In this section we give the description of all of the datasets used in this paper as well as the exact preprocessing steps.
\subsection{UCI ML datasets}\label{app:uci}
The Boston Housing dataset contains $506$ objects with $13$ features, the Abalone dataset has $4177$ objects with $8$ features and the Auto-mpg dataset consists of $398$ objects with $8$ features. For the Abalone dataset we add an additional feature by binning the ``diameter'' feature into $10$ bins uniformly distributed on $[0.1, 0.2]$ interval. We use this additional feature as constraints for evaluation of CLR-c approach (we also keep it for all other methods to provide better comparison). For the Auto-mpg dataset we split the categorical ``origin'' feature into $3$ binary features and we ignore the ``car name'' feature since it is unique for each object. We use model year as the constraints for the evaluation of the CLR-c approach. For the Boston Housing dataset we do not perform any additional preprocessing and use ``RAD''\footnote{Index of accessibility to radial highways.} feature as the constraints. To provide better comparison with results presented in k-plane regression paper~\cite{manwani2015k} we perform the same final preprocessing step by scaling all features into $[-1, 1]$ interval.

\subsection{Health insurance dataset}\label{app:health}
The health insurance dataset used in this paper consists of medical claims of patients from a particular health insurance provider. In total there are around $1$ billion claims, each characterized with $87$ different fields ($\approx 140$ Gb of the raw data). As a preprocessing step we keep only inbound (registered in a hospital), non-empty claims for $2014$ year. We delete claims containing mistakes, such as claims that have multiple patients associated with them, patients that have multiple gender or birthday or claims with total length of stay $\ge 60$ days. We process the data to obtain the following set of features. Numerical: age, DRG\footnote{Diagnosis Related Group~\cite{fetter1986diagnosis}.} weight, mean historic length of stay per DRG, mean historic length of stay per hospital, mean historic length of stay overall. Categorical: DRG, month of admission, hospital zipcode, was it observation stay, type of stay (emergency, urgent, elective, newborn or trauma), whether patient has been in this hospital before, whether patient had this DRG before. After dropping out claims with missing values and representing categorical features in one-hot encoding we obtain a corpus with $\approx 400000$ claims with $872$ features each. In order to further reduce the dimensionality we drop all the binary features that have value of $1$ in less than a $1000$ claims (e.g. certain rare DRGs or zipcodes). After the final preprocessing step the data size is still quite big $\approx 400000 \times 146$, however $94\%$ of all feature values are zeros. We consider length of stay as target features and use hospital membership as constraints for CLR-c approach.

\section{Experimental setup}\label{app:expsetup}
\subsection{Grid search parameters for UCI ML datasets}
For the three UCI ML datasets we evaluated the following regression algorithms:
\begin{itemize}
    \item \textbf{Linear regression (LR)}
    \item \textbf{Support vector regression (SVR)}; grid search over: \\$C \in \cbr{0.1, 1, 16, 32, 100, 128}, \sigma \in \cbr{1/d, 0.25, 0.5, 1}, \epsilon \in \cbr{2^{-8}, 0.01, 0.25, 0.5}$
    \item \textbf{Random forest (RF)}; grid search over: max depth $\in \cbr{\text{none}, 10, 50}$, max features $\in \cbr{d, 5}$, min samples split $\in \cbr{2, 10, 30}$, min samples leaf $\in \cbr{1, 10, 30}$, number of trees $= 30$.
    \item \textbf{k-plane regression}; grid search over: CLR $\in \cbr{\text{hard}, \text{soft}}$, weighted $\in \cbr{\text{dist}, \text{size}, \text{none}}$, $k \in \cbr{2, 4, 6, 8}$, $\gamma \in \cbr{0, 1, 10, 100}$, n-estimators $\in \cbr{1, 10}$
    \item \textbf{CLR-p}; grid search over: CLR $\in \cbr{\text{hard}, \text{soft}}$, weighted $\in \cbr{\text{true}, \text{false}}$, $k \in \cbr{2, 4, 6, 8}$, $\gamma \in \cbr{0, 1, 10, 100}$, n-estimators $\in \cbr{1, 10}$; random forest or logistic regression as label prediction model
    \item \textbf{CLR-c}; grid search over: $k \in \cbr{2, 4, 6, 8}$, $\gamma \in \cbr{0, 1, 10, 100}$, n-estimators $\in \cbr{1, 10}$    
\end{itemize}

For LR, SVR and RF we used scikit-learn implementations. The exact definition of the parameters can be found in the documentation of corresponding methods. The ``CLR'' parameter for the last three methods means which version of CLR was used to obtain clusters on the training data. The ``weighted'' parameter for k-plane indicate which weighting technique was used: ``none'' means original k-plane regression from~\cite{manwani2015k}, ``dist'' means the distance-based approach, ``size'' means the size-based approach of Bagirov et al.~\cite{bagirov2017prediction} both of which are described in section~\ref{seq:weight}. The parameters ``$k$'' and ``$\gamma$'' indicate the number of clusters and k-means regularization strength. ``n-estimators'' parameter greater than $1$ means that an ensemble method was used averaging the results from ``n-estimators'' different random CLR initializations. For CLR-p we either used random forest with $20$ decision trees and default other parameters or logistic regression with default $\alpha = 1.0$ as the label prediction model at test time. For all of the methods we limit the maximum number of CLR iterations to $5$ since in most cases it is enough to have good final prediction model. Also for stability reasons we use Ridge regression with regularization coefficient $\lambda = 10^{-5}$ instead of linear regression as basic regression model inside CLR.

Each algorithm is evaluated using 5 repetitions of 10-fold cross-validation. After the initial grid search for k-plane, CLR-p and CLR-c algorithms, we also tune the best models by changing underlying linear regression with Ridge regression and Lasso regression (with grid search over regularization parameter $\lambda \in \cbr{0.01, 0.1, 1, 10, 100}$). 

\subsection{Grid search parameters for medical claims dataset}
Since the health insurance dataset used in this paper is quite big, we run only one cross-validation with $3$ folds. Moreover, we only used hard version of CLR and ran it once for each $k$ and $\gamma$ per every fold of the cross-validation and shared the found labels across k-plane and CLR-p algorithms. We ran grid search over $k \in \cbr{2, 4, 6, 8}$ and $\gamma \in \cbr{0, 1, 10, 100, 1000}$, additionally checking the extremely big value of $\gamma = 10^5$ to confirm that k-plane regression works better by ignoring the CLR objective and only using k-means labels. We evaluated both weighted and not-weighted versions of all algorithms. As baselines we used only random forest and sparse linear regression (LSQR algorithm~\cite{paige1982lsqr}), since running SVR on this data turned out to be infeasible. For the random forest we used the following grid search parameters: max depth (md) $\in \cbr{\text{none}, 50}$, max features (mf) $\in \{\log_2 d, \sqrt{d}, d/5\}$, min samples split (mss) $\in \cbr{2, 10, 30, 50}$, min samples leaf (msl) $\in \cbr{1, 10, 30, 50}$. The number of trees was always $30$. For each of the best CLR-based models we additionally evaluated the ensemble of $10$ models with the same parameters. 

\subsection{Best parameters configuration}\label{app:best}
The best hyperparameters for the UCI ML experiments are the following:

\textbf{Boston housing}.

\begin{itemize}
    \item Random forest: max depth = 50, max features = 5, min samples split = 2, min samples leaf = 1
    \item SVR: $C = 128, \sigma=0.25, \epsilon=0.01$
    \item CLR-p RF (ens): $k = 8, \gamma = 10$, weighted = false, hard CLR with Lasso regression ($\lambda = 0.01$)
    \item CLR-p RF: $k = 6, \gamma = 10$, weighted = true, hard CLR
    \item CLR-p LR (ens): $k = 4, \gamma = 100$, weighted = false, hard CLR
    \item CLR-p LR: $k = 6, \gamma = 10$, weighted = true, hard CLR
    \item CLR-c (ens): $k = 6, \gamma = 0$
    \item CLR-c: $k = 6, \gamma = 10$
    \item k-plane (ens): $k = 6, \gamma = 100$, weighted = none, hard CLR with Ridge regression ($\lambda = 0.01$)
    \item k-plane: $k = 2, \gamma = 100$, weighted = none, hard CLR
\end{itemize}

\newpage
\textbf{Abalone}.

\begin{itemize}
    \item Random forest: max depth = 50, max features = all, min samples split = 30, min samples leaf = 10
    \item SVR: $C = 100, \sigma=0.25, \epsilon=0.5$
    \item CLR-p RF (ens): $k = 2, \gamma = 0$, weighted = true, soft CLR with Ridge regression ($\lambda = 0.1$)
    \item CLR-p RF: $k = 2, \gamma = 0$, weighted = true, hard CLR
    \item CLR-p LR (ens): $k = 8, \gamma = 100$, weighted = false, hard CLR
    \item CLR-p LR: $k = 6, \gamma = 100$, weighted = true, hard CLR
    \item CLR-c (ens): $k = 4, \gamma = 10$
    \item CLR-c: $k = 4, \gamma = 100$
    \item k-plane (ens): $k = 4, \gamma = 100$, weighted = none, hard CLR with Ridge regression ($\lambda = 0.1$)
    \item k-plane: $k = 4, \gamma = 100$, weighted = none, hard CLR
\end{itemize}

\textbf{Auto-mpg}.

\begin{itemize}
    \item Random forest: max depth = 10, max features = all, min samples split = 2, min samples leaf = 1
    \item SVR: $C = 32, \sigma=0.25, \epsilon=0.5$
    \item CLR-p RF (ens): $k = 8, \gamma = 1$, weighted = true, hard CLR with Lasso regression ($\lambda = 0.01$)
    \item CLR-p RF: $k = 6, \gamma = 1$, weighted = true, hard CLR
    \item CLR-p LR (ens): $k = 8, \gamma = 10$, weighted = false, hard CLR
    \item CLR-p LR: $k = 4, \gamma = 10$, weighted = true, hard CLR
    \item CLR-c (ens): $k = 4, \gamma = 100$
    \item CLR-c: $k = 2, \gamma = 100$
    \item k-plane (ens): $k = 8, \gamma = 10$, weighted = none, hard CLR with Lasso regression ($\lambda = 0.01$)
    \item k-plane: $k = 2, \gamma = 100$, weighted = none, hard CLR
\end{itemize}

\textbf{Medical claims}.

\begin{itemize}
    \item Random forest: n-est=30, max depth = none, max features = $d/4$, min samples split = 30, min samples leaf = 1
    \item CLR-p RF: $k = 8, \gamma = 1$, weighted = true, hard CLR
    \item CLR-p LR: $k = 4, \gamma = 10^5$, weighted = true, hard CLR
    \item CLR-c: $k = 8, \gamma = 0$
    \item k-plane: $k = 8, \gamma = 10^5$, weighted = none, hard CLR (ran for $20$ iterations)
\end{itemize}

All ensembles were evaluated with the same parameters as the best single methods with number of estimators = 10.

\subsection{Discussion of the best parameters}\label{app:disc}

\textbf{k-plane}. On average, the best parameters are not-weighted, hard CLR version of k-plane regression with as high $\gamma$ as possible and small number of clusters $2$ or $4$. Using ``size'' mode of weighting made the model consistently worse than linear regression on all datasets (which is reasonable, given that the final prediction is confined to be linear). Using ``dist'' mode of weighting made predictions better, when $\gamma$ is very small, since in that case k-means term does not affect the CLR algorithm and thus label prediction rule based on the distance to cluster centers is unjustified. Weighting the prediction makes the model more robust in that case. Using the soft version of CLR, k-plane regression shows somewhat worse results especially, when it is combined with weighting by distance. However, for the large $\gamma$ EM-based approach seems to be more stable, giving better average MSE. In general, k-plane algorithm would often produce extremely bad predictions resulting in very high MSE. The tendency to work better with the increase of the $\gamma$ is also concerning since it means that most of the gain comes from the use of k-means algorithm and not CLR, which means that the algorithm might perform as well by fitting a separate linear regression to the usual k-means clusters. This is especially visible in the results of k-plane regression applied to health insurance data.

\textbf{CLR-p}. On average, the best parameters are weighted, hard CLR version of CLR-p with medium $\gamma = 1$ or $10$. The number of clusters is less important since the algorithm shows good performance across all evaluated parameters of $k$. Weighted CLR-p is consistently better than non-weighted approach across all datasets, indicating a good quality of random forest probability predictions for CLR labels. Hard CLR is consistently better than soft CLR when $\gamma > 0$ and is usually worse when $\gamma = 0$. Logistic regression models do not show good accuracy which indicates the need of a flexible label prediction model.

\textbf{CLR-c}. The algorithm does not depend significantly on either $\gamma$ or $k$, although small number of clusters seem to work better for UCI datasets. On the other hand, for the health insurance experiments, the best results were obtained with $8$ clusters.

As expected, using ensembles with different CLR initializations work consistently better than single models across all methods and datasets. However, the bigger is the number of clusters, the larger is the improvement of using ensembles. Indeed, in that case there is more uncertainty in CLR algorithm and thus the solutions from different initializations have bigger variance. On average, using ensembles seems to decrease both the mean value of the error, as well as its standard deviation.

Finally, using Lasso and Ridge regressions instead of the standard linear regression inside CLR would often improve results even further. For example, the best results of CLR-p on Boston housing are obtained with Ridge regression and on Auto-mpg and Abalone with Lasso.

\section{EM equations for soft CLR}\label{app:em}

The soft CLR algorithm consists of two iteratively repeated steps.

\textbf{E-step}. Compute cluster membership probabilities with parameters $\cbr{\pi_i^{(t)}, \mu_i^{(t)}, w_i^{(t)}, b_i^{(t)}, \sigma_i^{(t)}}_{i=1}^k$:
\begin{equation*}
    q_{ji} = \frac{\pi^{(t)}_i\exp\rbr{-\frac{1}{2{\sigma_i^2}^{(t)}}\sbr{\rbr{y_j - x_j^Tw_i^{(t)} - b_i^{(t)}}^2 + \gamma\nbr{x_j - \mu_i^{(t)}}_2^2}}}{\sum_{c=1}^k\pi^{(t)}_c\exp\rbr{-\frac{1}{2{\sigma_i^2}^{(t)}}\sbr{\rbr{y_j - x_j^Tw_c^{(t)} - b_c^{(t)}}^2 + \gamma\nbr{x_j - \mu_c^{(t)}}_2^2}}}
\end{equation*}
\textbf{M-step}. Recompute the mixture parameters:
\begin{align*}
    \pi_i^{(t+1)} &= \frac{1}{n}\sum_{j=1}^nq_{ji},\
    \mu_i^{(t+1)} = \frac{\sum_{j=1}^nx_jq_{ji}}{\sum_{j=1}^nq_{ji}} \\
    w_i^{(t+1)}, b_i^{(t+1)} &= \argmin_{w, b}\sbr{\sum_{j=1}^nq_{ji}\rbr{y_j - x_j^Tw - b}^2} \\
    {\sigma_i^2}^{(t+1)} &= \frac{\sum_{j=1}^nq_{ji}\sbr{\rbr{y_j - x_j^Tw^{(t+1)}_j - b^{(t+1)}_j}^2 + \gamma\nbr{x_j - \mu_i^{(t+1)}}_2^2}}{\sum_{j=1}^nq_{ji}}
\end{align*}
We refer the reader to~\cite{desarbo1988maximum},~\cite{manwani2015k} for the derivation of these equations.

\newpage
\section{Dependence on the parameters for UCI ML datasets}\label{app:dep}

\begin{figure}[ht]
    \centering
    \includegraphics[width=1.0\textwidth]{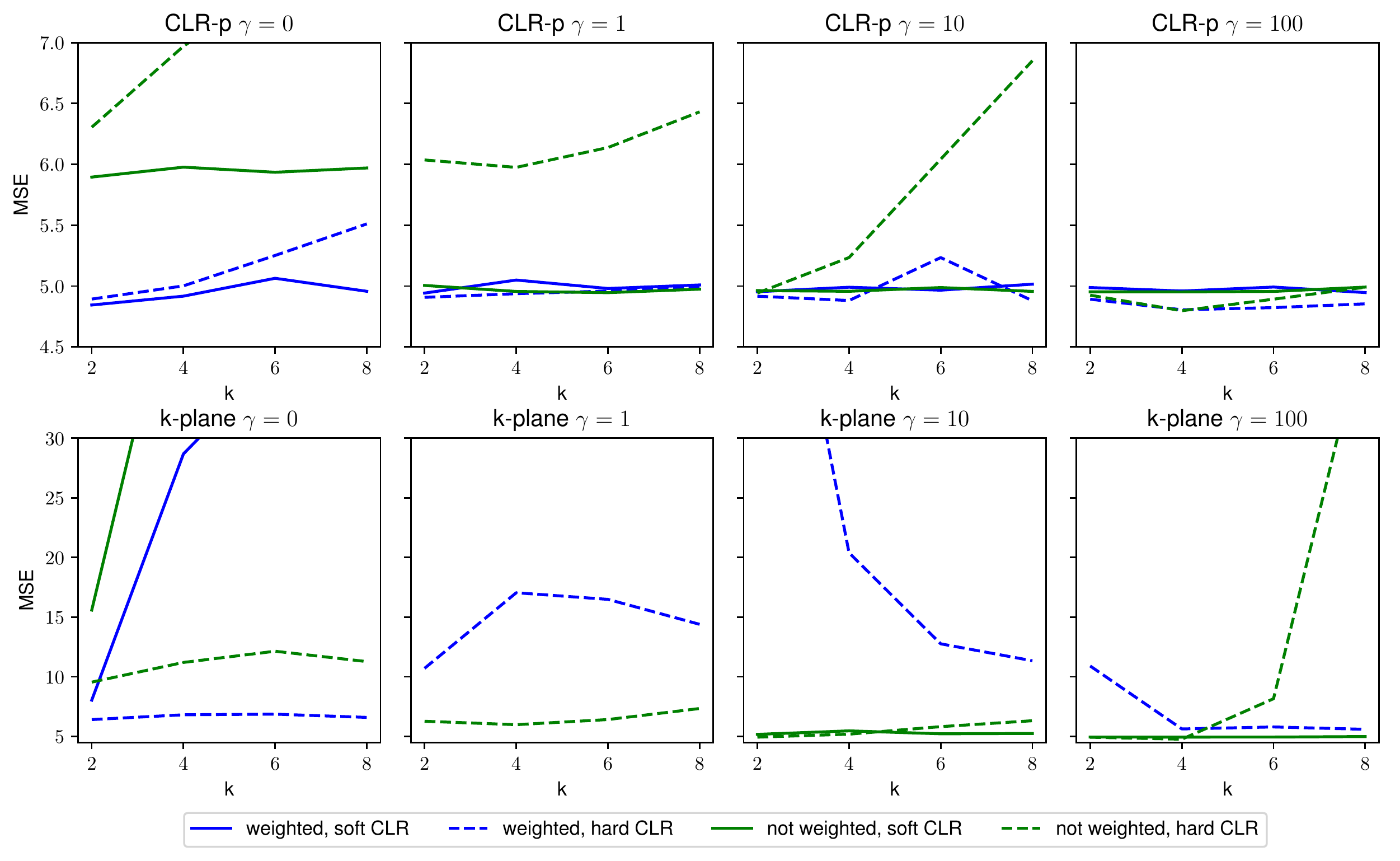}
    \caption{Abalone dataset, dependence on parameters for CLR-p and k-plane regression. $k$ is the number of clusters.}
    \label{fig:params-abalone}
\end{figure}

\begin{figure}[ht]
    \centering
    \includegraphics[width=1.0\textwidth]{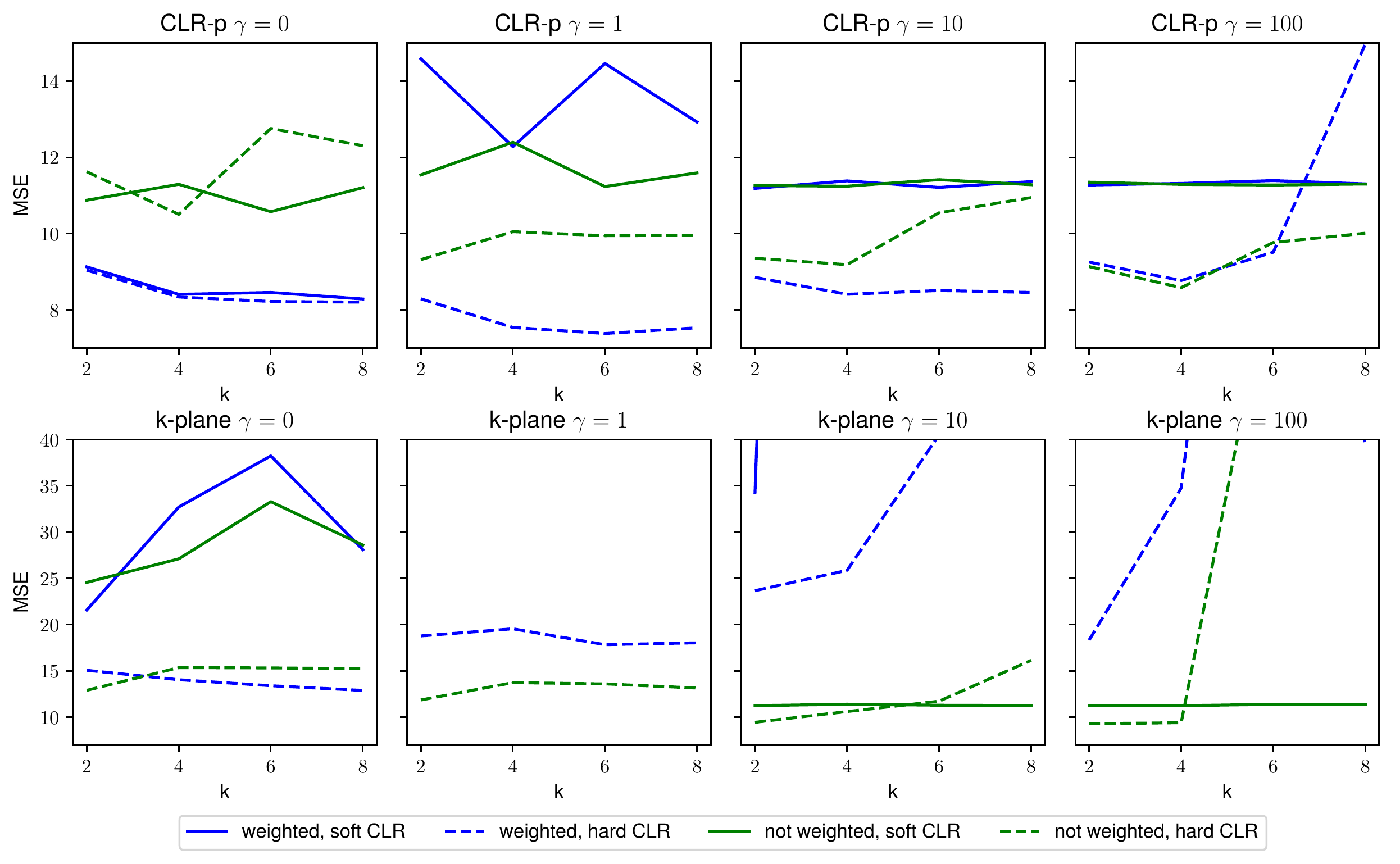}
    \caption{Auto-mpg dataset, dependence on parameters for CLR-p and k-plane regression. $k$ is the number of clusters.}
    \label{fig:params-autompg}
\end{figure}

\newpage
\begin{figure}[t]
    \centering
    \begin{subfigure}{0.49\textwidth}
        \centering
        \includegraphics[width=1.0\textwidth]{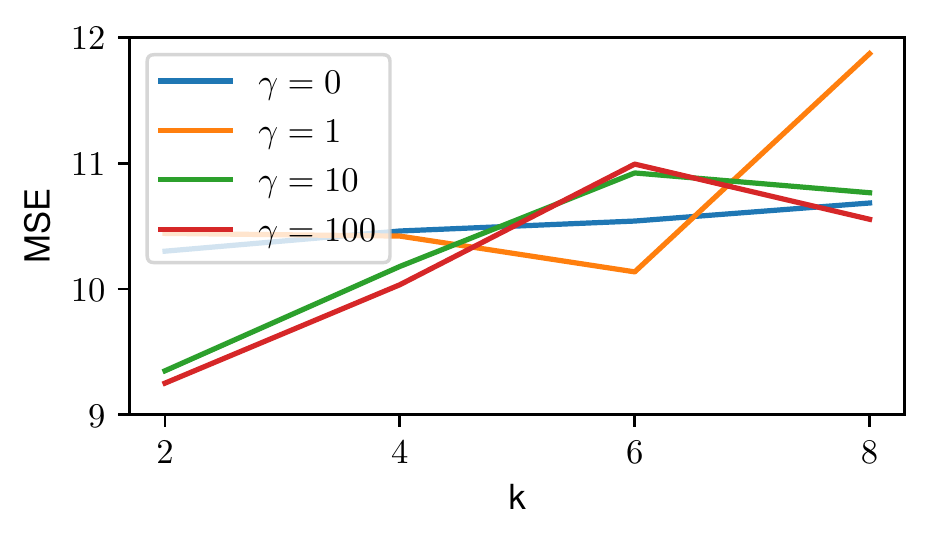}
        \caption{Auto-mpg}
    \end{subfigure}
    \begin{subfigure}{0.49\textwidth}
        \centering
        \includegraphics[width=1.0\textwidth]{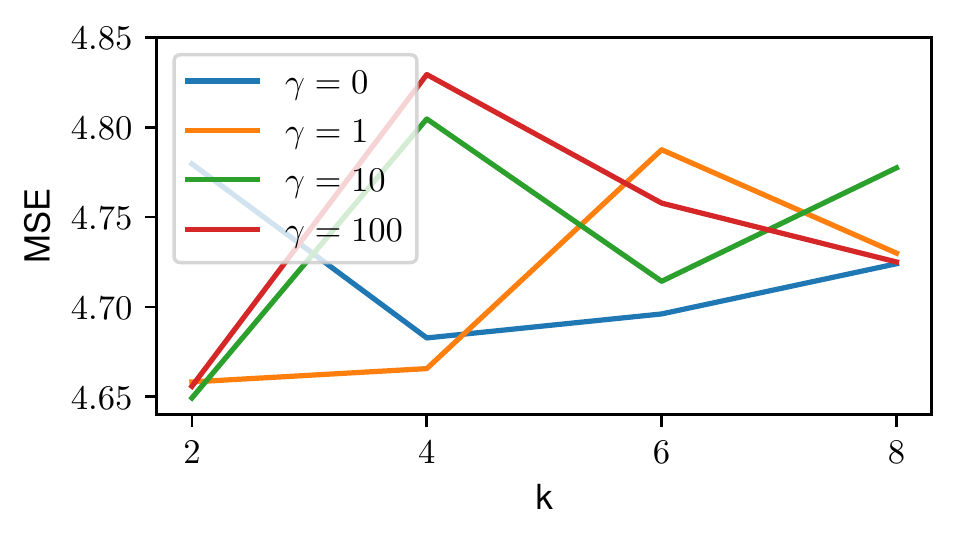}
        \caption{Abalone}
    \end{subfigure}
    \caption{CLR-c dependence on parameters. $k$ is the number of clusters.}
    \label{fig:CLRc_pm}
\end{figure}
\hbox{}

\end{document}